\pdfoutput=1

\documentclass[11pt]{article}

\usepackage[]{EMNLP2023}

\usepackage{times}
\usepackage{latexsym}
\usepackage{xcolor}

\definecolor{purp}{HTML}{791f87}

\makeatletter
\renewcommand{\sectionautorefname}{\S\@gobble}
\renewcommand{\sectionautorefname}{\S\@gobble}
\renewcommand{\subsectionautorefname}{\S\@gobble}
\renewcommand{\sectionautorefname}{\S\@gobble}
\renewcommand{\subsectionautorefname}{\S\@gobble}
\makeatother

\newcommand{\lm}[1]{\texttt{#1}}
\newcommand{\datasetname}{\mbox{\textsc{Tide}}\xspace}

\usepackage{color, colortbl}
\definecolor{azure}{rgb}{0.0, 0.5, 1.0}

\usepackage[T1]{fontenc}

\usepackage[utf8]{inputenc}

\usepackage{microtype}

\usepackage{inconsolata}

\usepackage[T1]{fontenc}

\usepackage{tabularx}
\usepackage{multirow}
\usepackage{rotating}

\definecolor{figurative}{HTML}{B2A9BC}
\definecolor{ambiguous}{HTML}{f7b2b1}
\definecolor{literal}{HTML}{f2ceb5} 
\definecolor{nonsense}{HTML}{c2cbcc}
\definecolor{sensitive}{HTML}{b2dcd2}
\definecolor{lm}{HTML}{A7C4D8}
\definecolor{nmt}{HTML}{EFB78F}

\title{\textit{That was the last straw, we need more:} \\
Are Translation Systems Sensitive to Disambiguating Context?}

\newcommand{\aspace}{\hspace{1em}}
\newcommand{\uw}{$^{\heartsuit}$}
\newcommand{\aiTwo}{$^{\clubsuit}$}

\author{
    Jaechan Lee\uw\aspace
    Alisa Liu\uw\aspace
    Orevaoghene Ahia\uw\aspace
    \textbf{Hila Gonen}\uw\aspace
    \textbf{Noah A. Smith}\uw\aiTwo\aspace\\
    \uw Paul G.\ Allen School of Computer Science \& Engineering, University of Washington \\
    \aiTwo Allen Institute for AI \\ \texttt{\{chan0369,alisaliu,oahia,nasmith\}@cs.washington.edu,hilagnn@gmail.com}
}

\begin{document}
\maketitle
\begin{abstract}
The translation of ambiguous text presents a challenge for translation systems, as it requires using the surrounding context to disambiguate the intended meaning as much as possible. 
While prior work has studied ambiguities that result from different \textit{grammatical} features of the source and target language, we study semantic ambiguities that exist in the source (English in this work) itself.
In particular, we focus on idioms that are open to both literal and figurative interpretations (e.g., \textit{goose egg}), and collect \datasetname,\footnote{Data and code can be found at \url{https://github.com/jaechan-repo/mt-ambiguity}.} a dataset of 512 pairs of English sentences containing idioms with disambiguating context such that one is literal (\textit{it laid a goose egg}) and another is figurative (\textit{they scored a goose egg}, as in a score of zero). 
In experiments, we compare MT-specific models and language models for (i) their \textbf{preference} when given an ambiguous subsentence, (ii) their \textbf{sensitivity} to disambiguating context, and (iii) the performance \textbf{disparity} between figurative and literal source sentences.
We find that current MT models consistently translate English idioms literally, even when the context suggests a figurative interpretation.
On the other hand, LMs are far more context-aware, although there remain disparities across target languages.
Our findings underline the potential of LMs as a strong backbone for context-aware translation.

\end{abstract}

\section{Introduction}

Natural language is inherently ambiguous due to the competing pressures of efficiency and clarity in communication \cite{zipf49,piantadosi-etal-2012}. 
As communicators, we disambiguate meanings on the basis of a wide range of contextual factors, or ask clarifying questions when such context is not available.
Though sometimes overlooked, the role of ambiguity in NLP has gained growing interest in recent work \cite{min-etal-2020-ambigqa,liu-etal-2023-afraid, stengeleskin-etal-2023-chicken}.

In machine translation (MT), it has long been recognized that ambiguities arise when the source language does not encode grammatical attributes that the target language requires \cite[i.a.]{barhillel-1953-some, prates-etal-2018-assessing, savoldi-etal-2021-gender, gonen-webster-2020-automatically}.
For instance, the English sentence ``\textit{I am a doctor}'' would require disambiguating the doctor's gender for translation to German, which has no gender-neutral word for ``\textit{doctor}.''
Prior work created contrastive test sets for such phenomena, to evaluate whether MT models correctly translate an ambiguous word (here, ``\textit{doctor}'') when disambiguating context is available (e.g., ``\textit{She is a doctor''})  \cite{muller-etal-2018-large, bawden-etal-2018-evaluating, voita-etal-2019-good}.

\begin{figure}[t]
    \centering
    \includegraphics[width=\columnwidth]{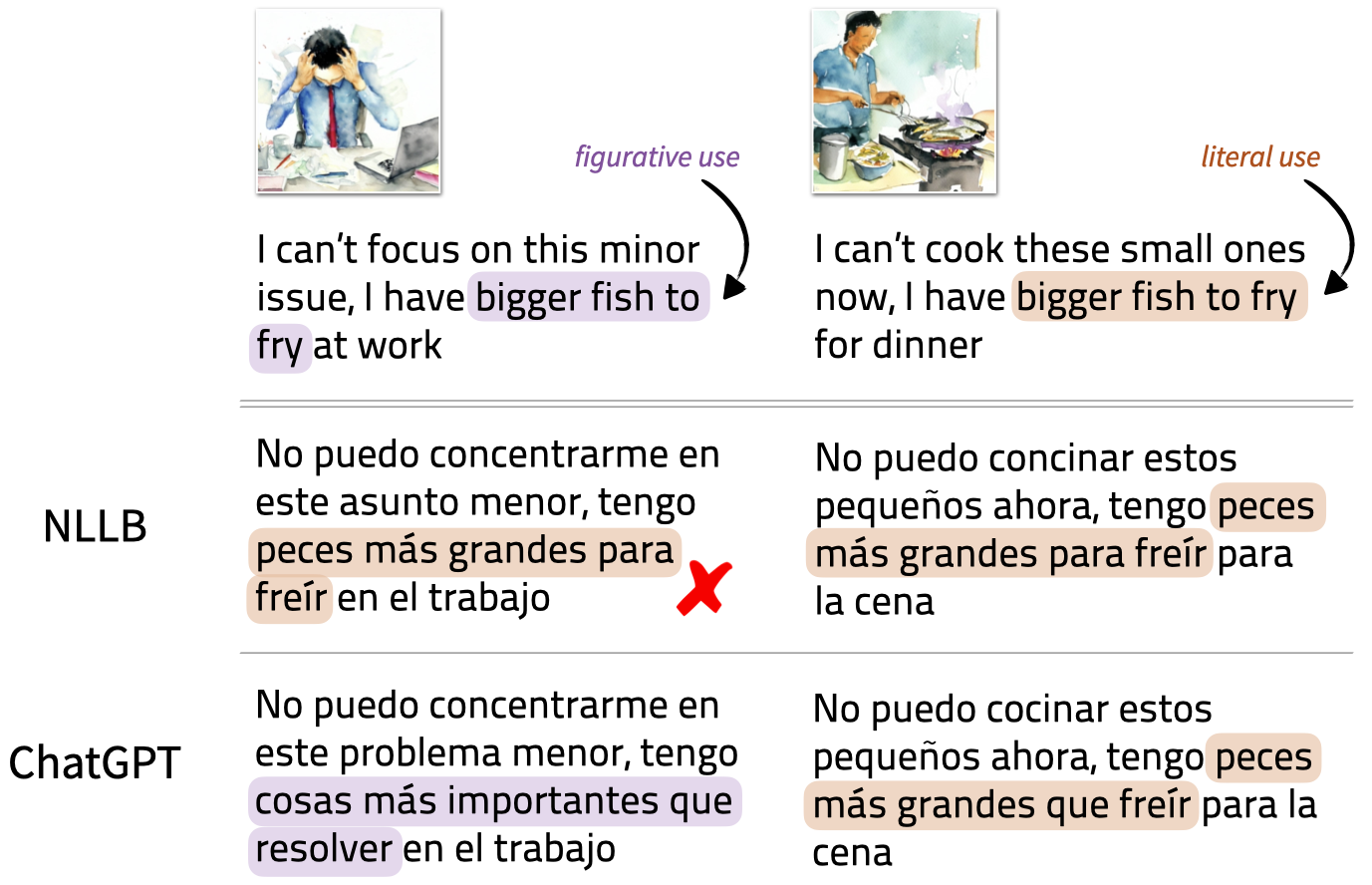}
    \caption{\datasetname consists of pairs of contrastive sentences that contain the same idiomatic expression in different contexts, such that one uses the figurative meaning of the idiom (left), and another uses its literal meaning (right). On this set of inputs, \lm{ChatGPT} is sensitive to the disambiguating context when translating the idiom, while \lm{NLLB} is not.}
    \label{fig:fig1}
\end{figure}

In contrast with \textit{grammatical ambiguity} with respect to a target language, it is relatively less understood how MT systems handle \textit{semantic ambiguity} present in the source text itself. 
For instance, \textit{``I have bigger fish to fry''} is ambiguous between figurative (\textit{``... at work''}) and literal (\textit{``... for the dinner''}) interpretations in English, outside of the context of translation. 
Therefore, we extend the line of work on context-aware translation to semantically ambiguous phrases in English.


To this end, we create \datasetname, \textbf{T}ranslations of \textbf{I}dioms in \textbf{D}isambiguating context in \textbf{E}nglish, a dataset of $512$ example triples.
Each triple consists of an ambiguous subsentence and a pair of contrastive sentences that contain the subsentence but add disambiguating context: one to produce a figurative interpretation of the idiom, and another to produce a literal interpretation of it (see \autoref{fig:fig1} for an example).
Our creation process for the triples combines automatic text generation with human annotation: we use \lm{GPT-4} to draft the triples, which are then scrutinized by human annotators. 
Following this, we engage native speakers from four languages to craft reference translations for a subset of the dataset.

In our experiments, we evaluate both traditional neural MT models and language models (LMs).
MT-specific models are trained on large corpora of parallel sentences, and have formed the foundation of translation research; LMs are trained without any explicit supervision for translation, yet recently demonstrate impressive translation ability \cite{hendy-etal-2023-how}.
Using \datasetname, we compare how these two types of systems handle ambiguity, and evaluate their sensitivity to disambiguating context. 
We find that on ambiguous input, LMs demonstrate roughly balanced preference between literal and figurative interpretations, whereas MT-specific models consistently prefer literal ones (\autoref{subsec:r1}).
Given disambiguating context, LMs are substantially more context-aware, though this sensitivity declines for more low-resource target languages; in contrast, MT-specific models tend to translate idioms literally irrespective of context (\autoref{subsec:r2}).
Finally, MT-specific models are better at translation of literal text than figurative text, whereas this disparity in LMs is much narrower (\autoref{subsec:r3}).

We summarize our contributions as follows: (1) We formalize the challenge of ambiguous idiomatic language in MT; (2) we create a new translation benchmark, \datasetname, that includes sentences with idioms along with disambiguating contexts (literal and figurative); (3) we analyze MT systems' behavior with and without disambiguating contexts, pointing to interesting trends and differences between LMs and MT-specific models.

\begin{table*}[t!]
    \centering
    \resizebox{\textwidth}{!}{%
    \begin{tabular}{lll}
        \toprule
        \textbf{Idiom} & \textbf{Figurative Sentence} & \textbf{Literal Sentence}\\\midrule
        \makecell*[{{p{4cm}}}]{tip of the iceberg\\\textit{to only know a very small part of the problem}} 
        & \makecell*[{{p{7cm}}}]{The problems we discovered were \textbf{just the tip of the iceberg} in this company.} 
        & \makecell*[{{p{7cm}}}]{As we approached the glacier, we saw \textbf{just the tip of the iceberg} above the water.} \\\midrule
        \makecell*[{{p{4cm}}}]{fall between the cracks\\\textit{be ignored or unobserved}}
        & \makecell*[{{p{7cm}}}]{His request for a promotion \textbf{fell between the cracks} due to the company's restructuring.}
        & \makecell*[{{p{7cm}}}]{The small toy \textbf{fell between the cracks} of the wooden floor.}\\\midrule
        \makecell*[{{p{4cm}}}]{foam at the mouth\\\textit{be extremely angry}}
        &\makecell*[{{p{7cm}}}]{He \textbf{was foaming at the mouth} when he found out about the betrayal.}
        &\makecell*[{{p{7cm}}}]{The rabid dog \textbf{was foaming at the mouth} and needed to be isolated.}\\\midrule
        \makecell*[{{p{4cm}}}]{foot in the door\\\textit{succeed with a first step}}
        &\makecell*[{{p{7cm}}}]{By volunteering at the company, she \textbf{got a foot in the door} for a full-time position.}
        &\makecell*[{{p{7cm}}}]{When the door was closing, he quickly \textbf{got a foot in the door} to prevent it from shutting.}\\
        \bottomrule
    \end{tabular}}
    \caption{\textbf{Examples in \datasetname}. A figurative and literal sentence disambiguates the idiom by adding context that demands figurative and literal interpretations, respectively.}
    \label{}
\end{table*}


\section{Creating \datasetname}

Idioms, though commonplace in daily communication, pose a challenge for MT systems due to its inherent ambiguity between literal and non-literal meanings. 
Generating the most appropriate translation among potential disambiguations of the idiom involves an understanding that extends beyond the idiom itself, as an MT system must use broader context clues to discern the most fitting translation.


We present \datasetname, a dataset of $512$ example triples. Each triple consists of an \textit{ambiguous subsentence}, a \textit{figurative sentence}, and a \textit{literal sentence} in English, all including the same idiom. 
The ambiguous subsentence permits both figurative and literal interpretations of the idiom, while the figurative and literal sentences introduce additional context that resolves the ambiguity to figurative and literal readings, respectively.
We design subsentences (e.g., \textit{``had a card up his sleeve''}) to be more than an idiom itself (here, \textit{``card up sleeve''}), as idioms alone can often be unnatural as standalone input to an MT system.

We construct \datasetname through a human-AI collaborative approach following a line of recent work \cite{liu-etal-2022-wanli, chakrabarty-etal-2022-flute}.
We first manually select candidate idioms from two large idiom corpora (\autoref{subsec:collect_idioms}).
Next, we leverage the generative power of \lm{GPT-4} to efficiently produce diverse and high-quality text, by prompting it to write a complete triple for each idiom (\autoref{subsec:generate_sentences}). 
To ensure quality and correctness, we then involve human annotators to filter out invalid triples (\autoref{subsec:crowdworker_labeling}).
Finally, we collect gold translations for a subset of the dataset among native speakers (\autoref{subsec:collecting_translations}).

\subsection{Collection of Idioms}\label{subsec:collect_idioms}
To collect idioms, we scrape \textsc{The Idioms} dictionary\footnote{\url{https://www.theidioms.com/}} to obtain 1409 idioms, and additionally use a dataset of 905 idioms from \citet{rabinovich2020pick}; both sources contain corresponding idiom definitions.
We discard duplicate idioms (including those that appear in different conjugations) and proverbs (e.g., \textit{All that glitters is not gold}), which are often too self-contained to be disambiguated with context.
Then, we manually select idioms that are available to a natural and plausible \textit{literal} interpretation, in addition to their figurative meanings.
This results in a set of 700 idioms with definitions.

\subsection{Generation of Idioms in Context}\label{subsec:generate_sentences}
Next, we draft an example triple for each idiom by prompting \lm{GPT-4} with a fixed prompt, containing two in-context examples along with additional guidelines (details in \autoref{subsec:appendix_generate_sentences}).
We write a set of heuristics to automatically identify some types of ill-formed output, such as when the subsentence is not an exact substring of the full sentences.
When a rule is violated, we add an additional turn of dialogue instructing the model to revise its output to follow the broken rule.
We repeat this until all rules are followed, or when two revisions are attempted without success.
After this, we have 700 English triples, each associated with a unique idiom.

\subsection{Human Annotation}\label{subsec:crowdworker_labeling}
Of course, the triples collected in \autoref{subsec:generate_sentences} may not correctly use idioms literally and figuratively, and generated text is susceptible to fluency and coherence issues.
To ensure data quality, we recruit crowdworkers on Amazon Mechanical Turk to label each of the full sentences as using either the literal or the figurative sense of an idiom.
We present each full sentence independently (not as a pair) to two different crowdworkers, who are asked to label it as \textit{figurative}, \textit{literal}, or \textit{ambiguous} with respect to how it uses the given idiom. 
They may also indicate that the sentence is invalid if it is offensive or has fluency issues (see \autoref{subsec:appendix_mturk} for details).

The annotators achieved substantial agreement on this task, with a Fleiss $\kappa$ score of 0.721. 
Furthermore, for 82.9\% of examples, there is a complete agreement between both annotators and the intended label (the label which we ask \lm{GPT-4} to follow when generating triples).

Based on the annotations, we discard triples where the intended-figurative sentences received no votes for figurative, or the intended-literal sentences received at least one vote not for literal. This asymmetry in the filtering heuristic is because we observe that \lm{GPT-4} was far more reliable at generating figurative uses of idioms than literal ones, and therefore we enforce a lower bar for retaining figurative sentences. We also discard all the triples that contain at least one vote for discard. In this way, we obtain the 512 English triples which constitute \datasetname.

\subsection{Collecting Translations}\label{subsec:collecting_translations}

Finally, for a randomly subset of 50 idioms, we gather reference translations for the contrastive pairs of figurative and literal sentences from native speakers of Hebrew, Yoruba, Korean, and Chinese. 

\section{Experimental Setup}

In this section we outline the models (\autoref{subsec:models}) and languages (\autoref{subsec:languages}) we evaluate, our automatic metrics (\autoref{subsec:metrics}), and our setup for collecting human evaluations of generated translations (\autoref{subsec:human_evaluation}).

\subsection{Models}\label{subsec:models}

We evaluate two classes of translation systems:  MT-specific models and LMs. 
Here, the MT-specific models use an encoder-decoder architecture and are trained on large amounts of parallel data, whereas the LMs are decoder-only models trained to maximize likelihood (i.e., next-token prediction) on predominantly-English text.

\paragraph{MT-Specific Models}
We evaluate \lm{NLLB} \cite{nllb-2022-no} and \lm{Opus MT} \cite{tiedemann-thottingal-2020-opus, tiedemann-2020-tatoeba}.
\lm{NLLB} is trained on partially synthetic parallel data, and covers 202 languages.\footnote{\url{https://huggingface.co/facebook/nllb-200-3.3B}}
\lm{Opus MT} is a collection of models, each with a fixed source and target language.\footnote{The most recent model for each language pair was downloaded from \url{https://github.com/Helsinki-NLP/Tatoeba-Challenge/tree/master/models}: transformer-big for $\texttt{De, Es,}$ and $\texttt{Hu}$, transformer-align for $\texttt{He}, \texttt{Hi}$, and $\texttt{Yo}$. 
 Their most recent English to Chinese models by July 2023 do not produce coherent outputs, so we proceed with the earlier version available on HuggingFace: \url{https://huggingface.co/Helsinki-NLP/opus-mt-en-zh}.
English to Korean models are not evaluated due to an issue with their PyTorch implementation, as reported by multiple users.}
For both models, we decode the translation greedily.

\paragraph{Language Models}
We evaluate \lm{ChatGPT} (\lm{gpt-3.5-turbo}; \citealt{openai-2022-chatgpt})\footnote{API last accessed on June 18, 2023.}
and \lm{PaLM 2} (\lm{text-bison-001}; \citealt{palm2}).\footnote{API last accessed on June 17, 2023.}
We do not include \lm{GPT-4} as it partially authored the examples in the dataset.

Both models were trained on a mixture of different languages, and in particular \lm{PaLM 2}'s training corpus included parallel data for hundreds of languages.
However, both LMs are trained for the next-token-prediction objective.

We prompt the LM to generate translations zero-shot with the prompt ``\texttt{Translate the following English sentence to [target language]: [source sentence]},'' and greedily decode the continuation.
We do not provide in-context examples or further instructions about figurative language, in order to create a setting comparable to the evaluation of MT-specific models.

\paragraph{Google Translate} 
We also include Google Translate\footnote{\url{https://translate.google.com/}. API last accessed on June 14, 2023.} for reference due to its popularity in commercial use. 
We do not classify it as either an MT-specific model or LM due to the lack of public understanding of how it works.

\subsection{Languages}\label{subsec:languages}
We consider the eight target languages: Spanish (\lm{Es}), Hindi (\lm{Hi}), German (\lm{De}), Hungarian (\lm{Hu}), Korean (\lm{Ko}), Chinese (\lm{Zh}), Hebrew (\lm{He}), and Yoruba (\lm{Yo}), which vary in resource-availability and are typologically and culturally diverse. 
When the evaluation requires a gold translation, we focus on the last four languages for which \datasetname contains human-written references.

\subsection{Automatic Metrics}\label{subsec:metrics}
We use different sets of metrics to evaluate translations for their literalness and for the overall translation quality.

\paragraph{Literalness} Following \citet{hendy-etal-2023-how}, we use two metrics to assess the literalness of the translation:
(1) \textit{Unaligned Source Words} (\lm{USW}) represents the number of source words unaligned with words in the translation, and (2) \textit{Non-Monotonicity} (\lm{NM}; \citealp{schioppa-etal-2021-controlling}) determines the extent of reordering in the word-to-word alignments from the source sentence to its translation.
For both metrics, we use the bitext alignments from the \texttt{awesome-align} framework \citep{dou2021word} which extract word alignments from \lm{mBERT} embeddings.

\paragraph{Translation quality}
We evaluate translation quality based on sentence similarity between reference and predicted translations.
We use \lm{chrF} \citep{popovic-2015-chrf}, \lm{BERTScore} \citep{sun-etal-2022-bertscore}, and \lm{BLEURT} \cite{sellam2020bleurt}.
\lm{chrF} measures precision, recall, and F-score of character $n$-grams.
\lm{BERTScore} is a contextual embedding-based evaluation metric that leverages the pretrained language model.\footnote{We use \lm{XLM-RoBERTa-base} embeddings for BERTScore. \citep{conneau2020unsupervised}} 
\lm{BLEURT} is a learned regression metric for automatic evaluation of generated text, which utilizes \lm{BERT} for training on pairwise comparisons of reference and candidate sentences, calibrated on human quality judgments.


\begin{figure*}
    \centering
    \includegraphics[width=\textwidth]{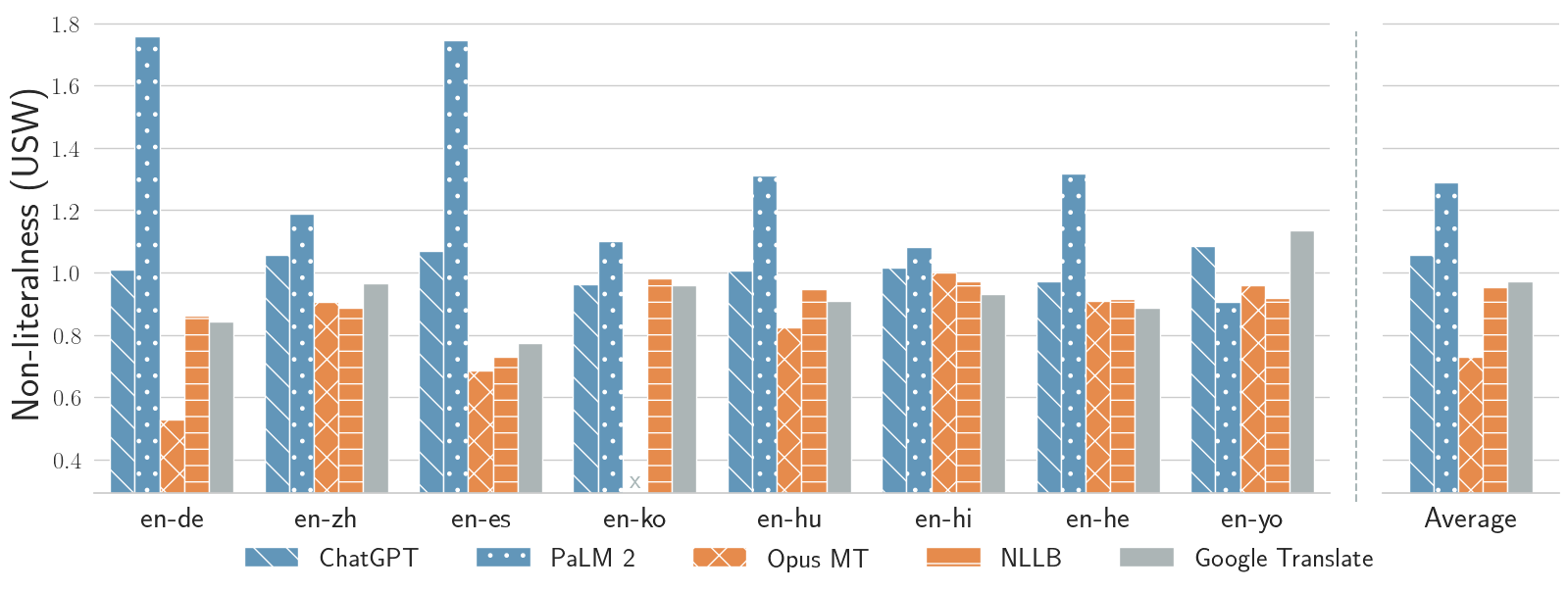}
    \caption{\textbf{Non-Literalness of Translations of Ambiguous Subsentences}, as measured by the number of \textit{unaligned source words} (USW) between the source sentence and its translation, normalized by the within-language average.
    Translations from \colorbox{lm}{pretrained LMs} are less literal than those of \colorbox{nmt}{MT-specific models}, suggesting that LMs prefer less literal translations of ambiguous input (i.e., without disambiguating context). En $\to$ Ko Opus MT models are not evaluated due to an issue with their implementation.}
    \label{fig:usw}
\end{figure*}

\begin{figure}[h]
    \centering
    \includegraphics[width=\columnwidth]{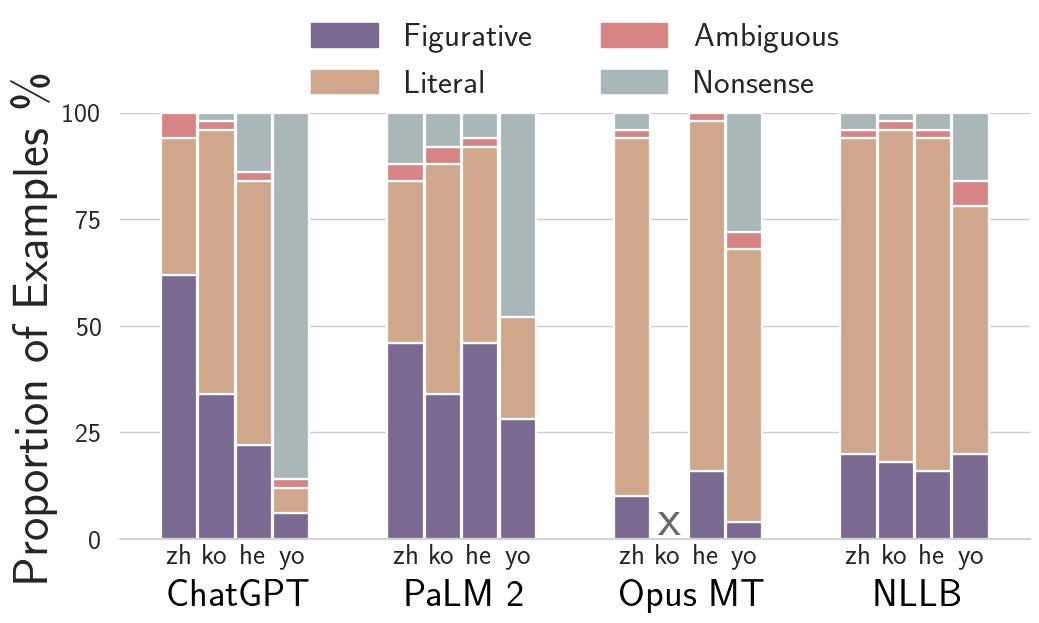}
    \caption{\small\textbf{Human Evaluation of Translations of Ambiguous Subsentences}, where annotators are asked to evaluate whether each translation is \colorbox{figurative}{figurative}, \colorbox{literal}{literal}, \colorbox{ambiguous}{ambiguous} due to an equivalent idiom, or is \colorbox{nonsense}{nonsensical}. \lm{ChatGPT} and \lm{PaLM 2} are more balanced in their preference between \colorbox{figurative}{figurative} and \colorbox{literal}{literal} translations; \lm{Opus} and \lm{NLLB} overwhelmingly prefer \colorbox{literal}{literal} translations.}
    \label{fig:r1_human}
\end{figure}

\subsection{Human Evaluation}\label{subsec:human_evaluation}

Due to the documented limitations of automatic evaluation for translation \cite{kasai-etal-2022-bidimensional}, we additionally perform human evaluation of model-generated translations for Chinese, Korean, Hebrew, and Yoruba.
We recruit one native speaker for each language, who are presented with the source sentences in each triple, along with generated translations from \lm{NLLB}, \lm{Opus MT}, \lm{ChatGPT}, and \lm{PaLM 2}.
The model-generated translations are presented in a random order not shown to the annotator.
For each sentence, they are asked: (1) Does the translation use the figurative meaning of the idiom, the literal meaning of the idiom, preserve the ambiguity due to an equivalent idiom in their language, or is it too nonsensical to determine? (2) Overall, is the translation perfectly correct, containing slight errors, or containing major errors?
We use the same subset of 50 triples from \autoref{subsec:collecting_translations}.
With 3 sentences per triple and 4 source models for each triple, annotators each evaluate $600$ translations.

\section{Experimental Results}\label{sec:experiments}

In our experiments, we explore MT-specific and LM systems' translation behavior on ambiguous subsentences (\autoref{subsec:r1}), their sensitivity to disambiguating context (\autoref{subsec:r2}), and their overall competence at translating literal versus figurative input (\autoref{subsec:r3}).

\begin{figure*}[t]
    \includegraphics[width=\textwidth]{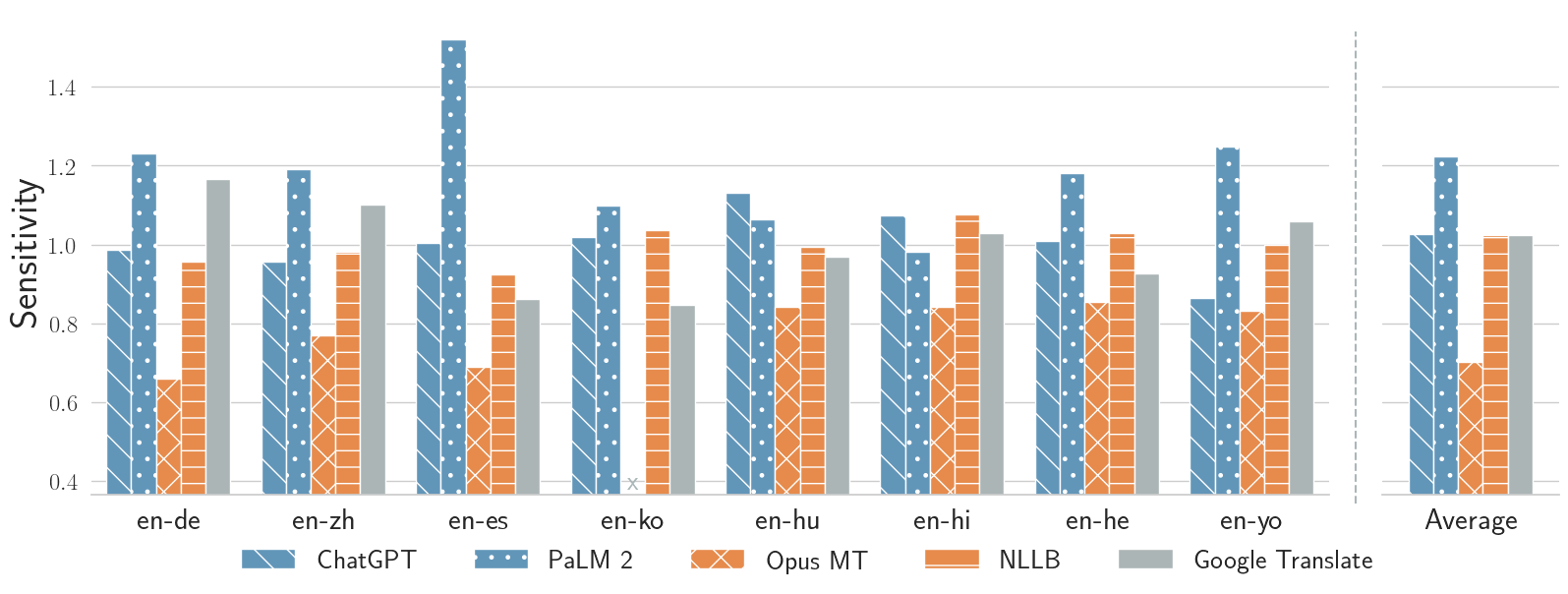}
    \caption{\textbf{Sensitivity to Disambiguating Context}, as measured by \lm{BERTScore-P}, describes how well an MT model adapts to disambiguating context for an otherwise ambiguous subsentence. 
    The metric is based on how the translation of the ambiguous subsentence changes between the two full sentences, and is normalized by the in-language mean.
    \colorbox{lm}{LMs} generally demonstrate greater context-awareness than \colorbox{nmt}{MT-specific models}.} 
    \label{fig:sensitivity}
\end{figure*}

\begin{figure}[h]
    \includegraphics[width=\columnwidth]{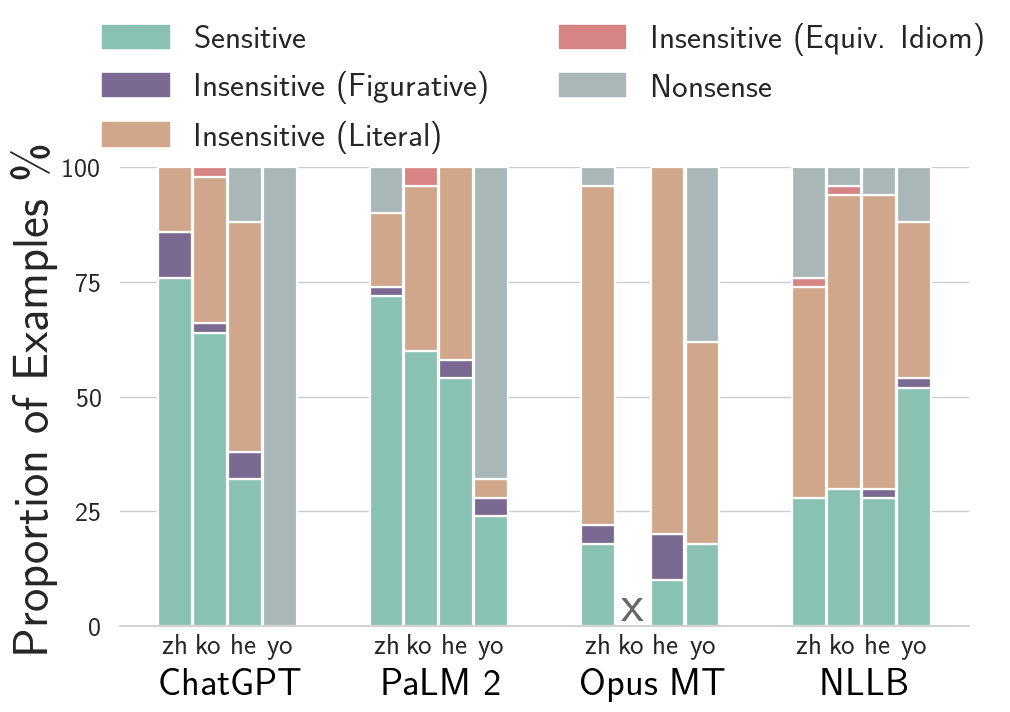}
    \caption{\small\textbf{Human Evaluation of Sensitivity}. A set of generated translations is considered \colorbox{sensitive}{context-sensitive}\xspace when it uses the figurative (or literal) sense of an idiom given figurative (or literal) disambiguating context. \lm{ChatGPT} and \lm{PaLM 2} are much more \colorbox{sensitive}{context-sensitive} than \lm{Opus MT} and \lm{NLLB}, which tend to translate idioms \colorbox{literal}{literally}\xspace irrespective of context.}
    \label{fig:rq2_human}
\end{figure}

\subsection{RQ1: How do MT systems translate ambiguous subsentences?}\label{subsec:r1}

First, we investigate how MT systems behave on ambiguous subsentences \textit{without} disambiguating context, in order to measure their preference for translating them figuratively or literally. 
We hypothesize that LMs are more likely to produce less literal translations of ambiguous subsentences than MT-specific systems, based on recent findings in \citet{raunak-etal-2023-gpts}.
Unlike their setting, here the source sentences are always ambiguous, so both literal and figurative translations are correct.



\paragraph{Automatic Evaluation} 
We measure the literalness of translations using \lm{USW} and \lm{NM}, where higher values mean less literal translations (\autoref{subsec:metrics}).
Within each language, we normalize values by the average across systems in that language.
This is because the metrics are not comparable across target languages, as they depend on linguistic properties of each target language.
Shown in \autoref{fig:usw}, LMs (in blue) produce translations with higher \lm{USW} scores than MT-specific models (in orange), across all target languages.
In particular, \lm{Opus MT} is the most literal model across all target languages. 
Moreover, we observe that the differences between LMs and MT-specific models become less pronounced for more under-resourced languages (the languages are ordered left to right based on count of pages in Common Crawl\footnote{\url{https://commoncrawl.github.io/cc-crawl-statistics/plots/languages}}).

Results based on \lm{NM} (shown in \autoref{sec:appendix_additional_results}) corroborate our findings for SVO languages. This metric is inherently limited to target languages with the same word order as the source language (English in this work, with SVO order).


\begin{figure*}[t]
    \includegraphics[width=\textwidth]{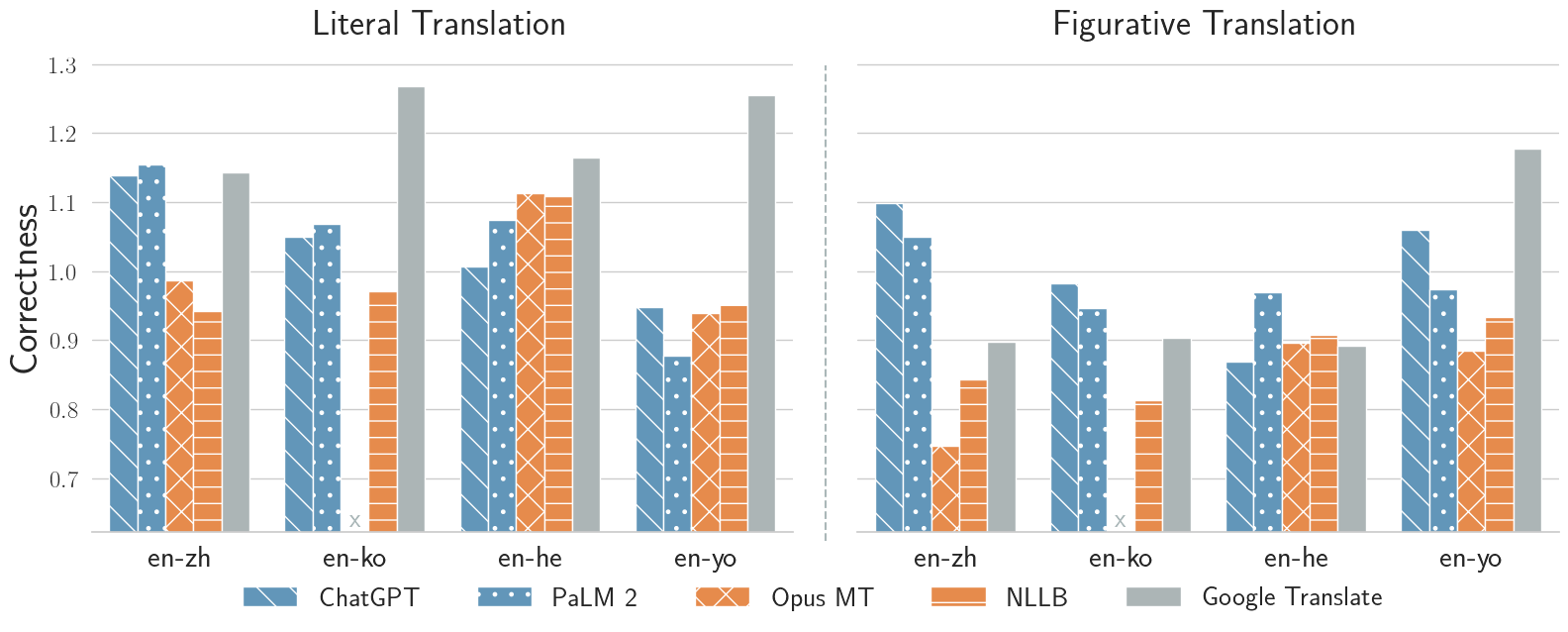}
    \caption{\textbf{Overall Translation Quality for Literal (left) and Figurative (right) Sentences}, as measured by BLEURT between the reference and prediction. 
    While \colorbox{lm}{LMs} and \colorbox{nmt}{MT-specific models} show comparable performance in translating literal sentences, NMT models are much weaker on figurative source sentences.}
    \label{fig:correctness}
\end{figure*}

\paragraph{Human Evaluation} 
In \autoref{fig:r1_human}, we show the human judgments of translations of ambiguous subsentences, indicating whether the translation is ambiguous, literal, figurative or nonsense.
These results corroborate findings from automatic evaluation, and show even clearer distinctions.
Overall, \lm{ChatGPT} and \lm{PaLM 2} demonstrate much more balanced preferences between figurative and literal translations, compared to \lm{Opus MT} and \lm{NLLB}.
For the target language Chinese, \lm{ChatGPT} prefers a figurative translation 62\% of the time; however, that preference declines dramatically as the target language becomes more low-resource, dropping to 6\% for Yoruba.
\lm{PaLM 2} demonstrates more robust preferences across target languages, consistently preferring figurative translations 28\% to 46\% of the time.
In contrast, \lm{Opus MT} and \lm{NLLB} overwhelmingly prefer literal translations, choosing a figurative translation only 4\% to 20\% of the time.

\subsection{RQ2: How sensitive are MT systems to disambiguating context?}\label{subsec:r2}

We next explore to what extent the predicted translation of an ambiguous subsentence changes when disambiguating context is available. 

\paragraph{Automatic Evaluation}

Intuitively, if the LM is not sensitive to context, then the translation of the ambiguous subsentence, $p_a$, should be equally contained in the translation $p_\ell$ for the literal sentence, and the translation $p_f$ for the figurative sentence.
That is, the way the ambiguous subsentence $a$ is translated should not be affected by the added context.
On the other hand, if $p_a$ is more contained in $p_\ell$ than in $p_f$ (or vice versa), that would mean how the model handles $a$ changes with the context.


Therefore, we operationalize the sensitivity to disambiguating context as
\begin{equation*}
    \lvert\texttt{contained\_in}(p_a, p_l) - \texttt{contained\_in}(p_a, p_f)\rvert
\end{equation*}
where $\texttt{contained\_in}()$ is a measure of unidirectional sentence similarity.
Here, we use \lm{chrP} and \lm{BERTScore-P}, the precision outputs of \lm{chrF} and \lm{BERTScore}, both ranging from 0 to 1.
A higher value of sensitivity (close to 1) indicates high sensitivity to disambiguating contexts.

\autoref{fig:sensitivity} shows the sensitivity results for the different models. The LMs, \lm{PaLM 2} and \lm{ChatGPT}, generally exhibit a higher degree of sensitivity across most language pairs. Comparatively, the MT-specific models, \lm{Opus MT} and \lm{NLLB}, show less sensitivity. \lm{Opus MT}, in particular, consistently demonstrates the lowest context sensitivity for all target languages.

\paragraph{Human Evaluation} 

In human evaluation, a model is considered context-sensitive on a triple if annotators indicate that the idiom is translated figuratively for the figurative sentence, and literally for the literal sentence.
Otherwise, the model is insensitive.
As shown in \autoref{fig:rq2_human}, both \lm{ChatGPT} and \lm{PaLM 2} are very sensitive to context, though there is still room for improvement.
For instance, for \lm{En}$\to$\lm{Zh} translation, \lm{ChatGPT} and \lm{PaLM 2} translate correctly for $76\%$ and $72\%$ of idioms, respectively.
Yet, the sensitivity of both models declines monotonically as the target language becomes more low-resource.
In particular, for \lm{En}$\to$\lm{Yo} translation, \lm{ChatGPT} translations are entirely nonsensical, and are qualitatively reported as frequently containing hallucinations completely unrelated to the source.

Nonetheless, \lm{Opus MT} and \lm{NLLB} are substantially less context-aware, correctly adapting to disambiguating context only $11.5\%$ and $34.5\%$ of the time, respectively.
Yet, their more consistent performance across languages suggests that dedicated training for translation leads to better results on low-resource languages.

\begin{figure*}[t]
    \centering
    \includegraphics[width=\textwidth]{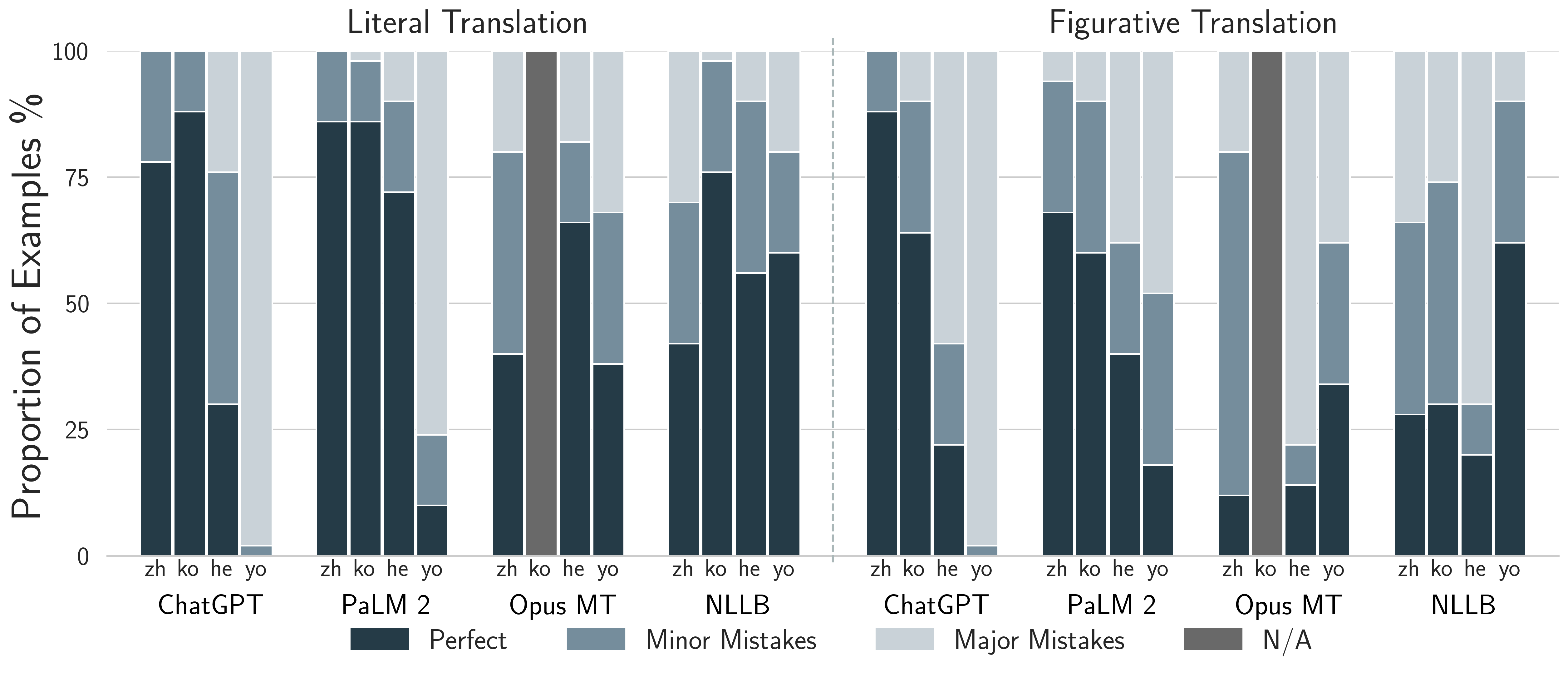}
    \caption{\textbf{Human Evaluation of Overall Translation Quality}, reported separately for figurative versus literal source sentences. \lm{Opus} and \lm{NLLB} are substantially better at literal translation than figurative translation overall, whereas \lm{ChatGPT} and \lm{PaLM 2} exhibit a much smaller disparity between literal and figurative translation quality.}
    \label{fig:rq3_human}
\end{figure*}

\subsection{RQ3: Are there performance disparities between figurative and literal translations?}\label{subsec:r3}

Finally, we investigate if translation systems have systematic performance gaps between translating figurative versus literal input.

\paragraph{Automatic Evaluation}
We use the reference translations collected in \autoref{subsec:collecting_translations}, and measure text similarity between predicted and reference translation with \lm{BLEURT}. 





The results are shown in \autoref{fig:correctness}.
Across the board, models are more capable at literal translation than figurative translation.
Yet, the gap is more pronounced for MT-specific models compared to LMs.
\lm{ChatGPT} and \lm{PaLM 2} exhibit performance gaps of $2.92\%$ and $4.85\%$, respectively, between literal (higher) and figurative translations, on average across languages.
For \lm{OPUS} and \lm{NLLB} this disparity is higher: $16.4\%$ and $11.7\%$, respectively.

Overall, MT-specific models and LMs demonstrate comparable performance on literal translations, while NMT models lag behind LMs on figurative translations.




\paragraph{Human Evaluation}
In \autoref{fig:rq3_human}, we compare how human annotators evaluate the correctness of translations overall, with the options \textit{perfect}, \textit{minor mistakes}, and \textit{major mistakes}.
Consistent with findings from automatic evaluation, \lm{ChatGPT} and \lm{PaLM 2} demonstrate more consistent performance across literal and figurative translations. 
However, \lm{Opus} and \lm{NLLB} are notably stronger at literal translations than figurative ones.

We additionally observe that on Yoruba, the most low-resource language we study, \lm{Opus MT} and \lm{NLLB} actually far outperform \lm{ChatGPT} and \lm{PaLM 2}.
We speculate that pretrained LMs are particularly strong on languages that were well-represented during pretraining; when this is not the case, it may produce degenerate text by entirely failing to grasp the translation task.

\section{Related Work}


\paragraph{Ambiguity in translation}
Context-aware translation usually focuses on grammatical features that the source language does not encode but the target language requires, such as formality (e.g., Chinese has a formal and informal ``\textit{you}''; \citealp{voita-etal-2019-context}), gendered pronouns (e.g., French has a male and female ``\textit{it}''; \citealp{muller-etal-2018-large, yin-etal-2021-context}), verb form (e.g., Spanish has six verb forms for past tense; \citealp{fernandes-etal-2023-translation}), and ellipses (e.g., ``\textit{We all did}'' in English cannot be translated to Russian without identifying the elided verb; \citealp{voita-etal-2019-good}).
Another well-studied issue is lexical cohesion, where the same phrase in the source sentence (e.g., a named entity like ``\textit{Julia}'') should be translated consistently each time \cite{wong-kit-2012-extending,kuang-etal-2018-modeling}.
In contrast, our work extends the study of context-aware translation to expressions which are ambiguous \textit{in the source language alone}, focusing on idiomatic expressions.
\datasetname joins a family of contrastive datasets that test model sensitivity to contextual information \cite[i.a.]{muller-etal-2018-large, bawden-etal-2018-evaluating, voita-etal-2019-good}.

\paragraph{Translation of figurative language} 
Figurative language has received considerable attention in MT research.
Some work has studied the hidden representations or attention patterns of MT-specific models when processing multi-word expressions \cite{rikters-bojar-2017-paying, garcia-etal-2021-probing, dankers-etal-2022-transformer}, or proposed methods to improve translation of these expressions \cite{zaninello-birch-2020-multiword, gamallo-garcia-2019-unsupervised}.
In particular, \citet{baziotis-etal-2023-automatic} show that monolingual pretraining improves figurative translation, which may explain our finding that pretrained LMs generate less literal translations and are more sensitive to disambiguating context.

The most closely related work, \citet{raunak-etal-2023-gpts}, compare how LMs and MT-specific systems translate sentences with idiomatic expressions, and similarly find that LMs produce substantially less literal translations.
We go further by evaluating how these models handle \textit{ambiguous} input and their \textit{sensitivity} to disambiguating context.

\paragraph{Datasets for idiom translation}
\citet{fadaee-etal-2018-examining} introduced the first extensive dataset for idiom translation, identifying data scarcity as one of core challenges in this domain.
EPIE \cite{saxena2020epie} is a large-scale corpus with 25K potentially idiomatic expressions (PIEs), with representation of both figurative and literal usages. MAGPIE \cite{haagsma-etal-2020-magpie} is a more expansive dataset of 50K samples that also contain genre labels. PECTI \cite{tang2022petci} curated a parallel English translation dataset of Chinese idioms.
While these datasets offer a general-purpose testbed, the contrastive sentence pairs in \datasetname enable finer-grained analysis, while the fluency of source sentences matches (if not exceeding) that of naturally-occurring datasets.

\section{Conclusion}

In this work we focus on semantic ambiguity in machine translation, specifically when using idiomatic language. We introduce a new benchmark (\datasetname) of sentences that include idioms, along with disambiguating contexts (both literal and figurative). We then use \datasetname to investigate the behavior of different translation systems on ambiguous input and their sensitivity to disambiguating context, uncovering new strengths of pretrained LMs compared to MT-specific models.

Our findings point to pretrained LMs as a promising backbone for translation systems, and we foresee a future that combines the strong language understanding of LMs with dedicated supervision for translation.

\section*{Acknowledgments}
We would like to thank the UW NLP community for valuable discussion of this work.
We are grateful to Weijia Shi, Jiacheng (Gary) Liu, and Xiaochuang Han for their help in writing and evaluating Chinese translations, and Zhaofeng Wu for feedback on the draft and figures.

We thank the reviewers for their valuable feedback and suggestions, and OpenAI for offering access to their models through the API.

\section*{Limitations}
In this work we study ambiguous source sentences specifically through idioms that are available to both literal and figurative interpretations.
While this allows us to efficiently collect a dataset and perform focused evaluation, ambiguity occurs in more diverse forms, and we encourage future work to collect more data in the form of \datasetname.
Contemporary work collects a dataset of ambiguous sentences (with direct disambiguations, rather than disambiguating context), and is a promising start \cite{liu-etal-2023-afraid}. 

In addition, we only study the behavior of translation systems when English is the source language, due to the availability of English idiom collections.
Yet figurative expressions vary greatly across languages \cite{kabra-etal-2023-multi}, and our conclusions may not necessarily generalize to translation from other languages.


\bibliography{anthology,custom}
\bibliographystyle{acl_natbib}

\appendix

\section{\datasetname Creation Details}\label{subsec:appendix_generate_sentences}

\subsection{Sentence generation}

We use \lm{GPT-4} to generate the 700 triples consisting of an ambiguous subsentence, a figurative sentence, and a literal sentence. 
The configuration parameters were set as follows: \texttt{max\_tokens=512}, \texttt{temperature=0}, and \texttt{top\_p=1}.
The prompt is shown in \autoref{tab:generate_data}. 

In addition, the generation process undergoes iterative refinements under a set of criteria, during which we prompt the \lm{GPT-4} instance to rewrite the entire triple if the generation included any prohibited words: ``\textit{literally}'', ``\textit{figuratively}'', ``\textit{ambiguously}'', ``\textit{physically}'', ``\textit{metaphorically}'', and ``\textit{because}''. 
These words are observed to potentially degrade sentence quality, as they often prompt the \lm{GPT-4} to merely provide working definitions of the idioms instead of generating novel context. 
We also ensure through these refinements that the ambiguous subsentence is indeed a substring of the figurative and literal sentences.

\subsection{Processing pronouns}

As written in the generation prompt (\autoref{tab:generate_data}), we ban \lm{GPT-4} from including subjects in the ambiguous subsentence as we observe that \lm{GPT-4} frequently uses personal pronouns which end up disambiguating the whole subsentence (e.g., \textit{He is a chip off the old block} is not ambiguous due to the pronoun \textit{he}).
Following the generation stage, we conduct additional rule-based modifications to the sentences to facilitate the translation process for MT models. 
In cases where the ambiguous subsentence begins with a lexical verb, and both the literal and figurative sentences include interchangeable subjects preceding the verb, we make alterations so that both use the same pronoun, which are then incorporated into the shared subsentence. These alterations include converting ``he'' to ``she'', ``she'' to ``he'', and ``he''/``she'' to ``they'' to have the pronoun shared between the figurative and literal sentence.

\begin{figure}
    \centering
    \includegraphics[width=0.9\linewidth]{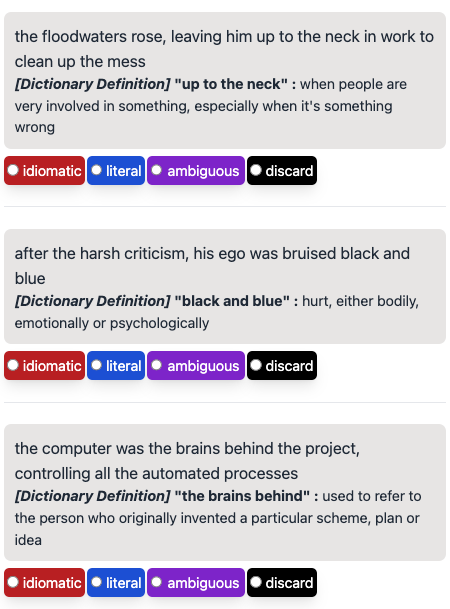}
    \small
    \caption{\textbf{Amazon Mechanical Turk (MTurk) Worker Interface}, containing 3 example problems.}
    \label{fig:interface}
\end{figure}

\section{Amazon Mechanical Turk (MTurk) details}\label{subsec:appendix_mturk}

We employ Amazon Mechanical Turk (MTurk), a crowdsourcing marketplace, to collect well-formed triples, composed of an idiom and corresponding ambiguous subsentence, figurative sentence, and literal sentence, generated by \lm{GPT-4} as described in \autoref{subsec:generate_sentences}.

We select 30 workers based on their scores in a qualification test human intelligence task (HIT) that we administer. This test, which typically requires less than 30 minutes to complete, consists of 20 handcrafted problems in the exact format as the main HIT. Upon completion, workers receive a payment of \$7.

In both the qualification test and the main task, each problem presents an English utterance derived from a randomly shuffled pool of 700 ambiguous subsentences, 700 figurative sentences, and 700 literal sentences. The problem also provides the corresponding idiom in use and its dictionary (figurative) definition. With this information, workers must ascertain whether the idiom was used in a \textit{figurative}, \textit{literal}, or \textit{ambiguous} context, or if the utterance should be \textit{discard}ed. The option to discard is included to eliminate nonsensical or offensive generations.

For the main task, we gather multiple gold labels for each problem to ensure accuracy and credibility. This means that multiple workers are assigned the same problem. We initially release a batch of first 50 problems of the pool, collecting 4 gold labels for each to examine interannotator agreement. Based on our observation that 2 gold labels are sufficient, we proceed to collect only 2 labels for the remaining batches. Workers are remunerated at a rate of \$0.30 for every 5 problems completed, an interval expected to take 1 minute.

For how we utilize these labels, see \autoref{subsec:crowdworker_labeling}.

\section{Additional results}\label{sec:appendix_additional_results}

\begin{figure*}
    \centering
    \includegraphics[width=\textwidth]{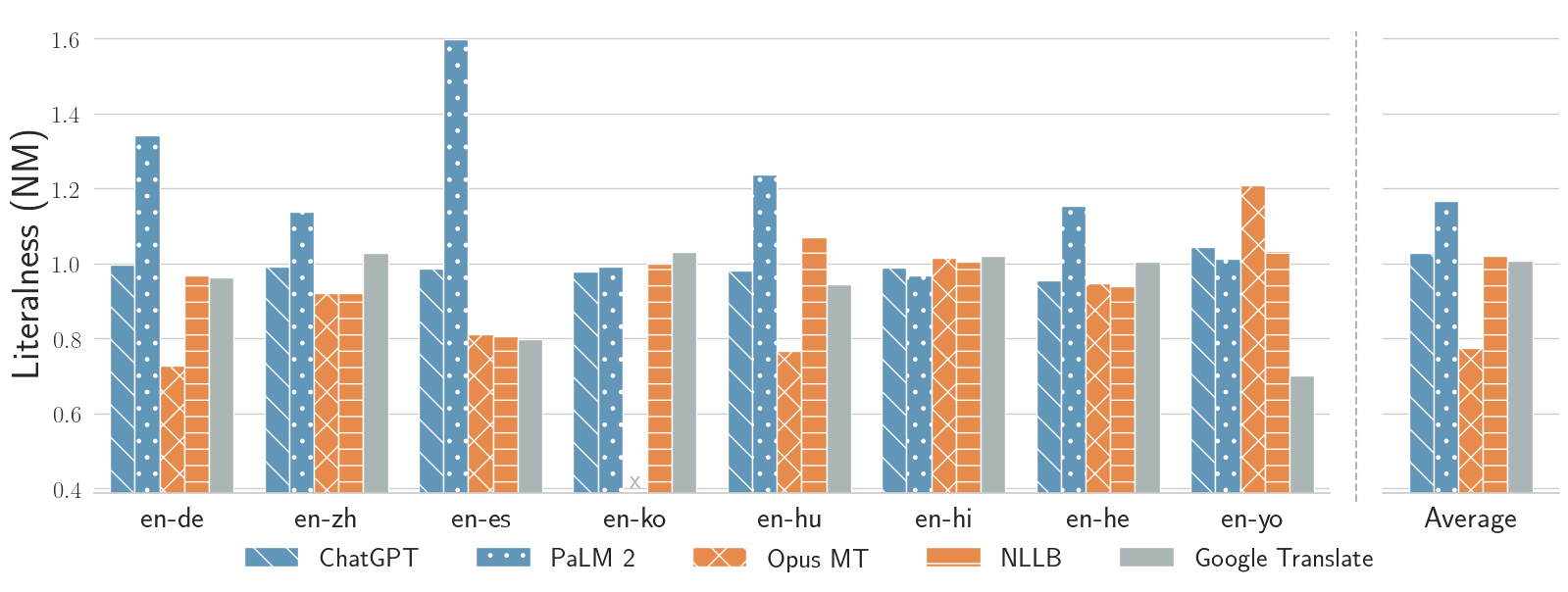}
    \caption{\textbf{Non-Literalness of Ambiguous Subsentences}. This graph visualizes non-monotonicity (NM) between the translation of each language pair, quantifying the level of word order shifts that occur during translation.}
    \label{fig:nm}
    \vspace{0.5cm}
    \includegraphics[width=\textwidth]{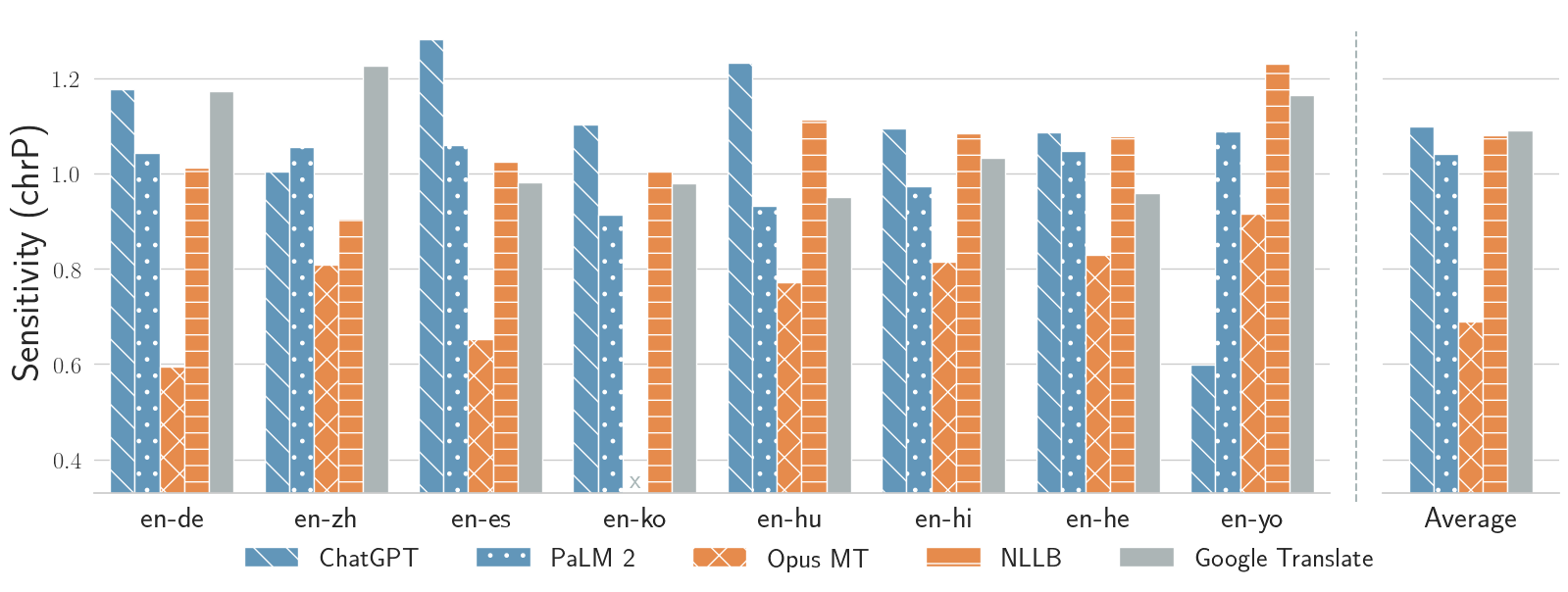}
    \caption{\textbf{Sensitivity}, $|\text{contained\_in}(p_a, p_f) - \text{contained\_in}(p_a, p_l)|$, computed using chrP (the precision score of chrF) as the implementation of contained\_in.}
    \label{fig:sensitivity_chrf}
\end{figure*}

See \autoref{fig:nm} and \autoref{fig:sensitivity_chrf} for additional results on RQ1 and RQ2, respectively.

\begin{table*}[t]
    \centering
    \begin{tabularx}{\textwidth}{lX}
        \toprule
        
        \multirow{1}{*}{\rotatebox[origin=c]{90}{\parbox[c]{5cm}{\large\textbf{System Instruction}}}}&\footnotesize
        \texttt{System}: Your task is to write an AMBIGUOUS phrase, a FIGURATIVE sentence, and a LITERAL sentence that use the given IDIOM. Consider the IDIOM ``burn the bridge'' as an example. An AMBIGUOUS phrase would allow for both figurative and literal interpretations of the IDIOM. For example, the phrase ``burned the bridge with him'' can be interpreted figuratively (in context of interpersonal relationships) or literally (burning down the physical bridge). A FIGURATIVE sentence must add extra words to your AMBIGUOUS phrase to only allow for a figurative, metaphorical interpretation. For example, the sentence ``She burned the bridge with him because he publicly criticized her works'' only makes sense in the figurative context of human relationships. A LITERAL sentence must add extra words to your AMBIGUOUS phrase to only allow for a literal, word-by-word interpretation. For example, the sentence ``She burned the bridge with him after crossing it'' only makes sense in literal context of a physical bridge that can be crossed.
        Note these warnings:
        \begin{itemize}
            \itemsep0em 
            \item Unless the idiom already contains a subject, your ambiguous phrase must not introduce any subjects.
            \item Do not end your ambiguous phrase with a preposition.
            \item Do not write more than one sentence for each figurative or literal sentence.
            \item Surround your AMBIGUOUS phrase with delimiters <>.
            \item When you write the LITERAL and FIGURATIVE sentences, do NOT modify the spelling of the AMBIGUOUS phrase in any way. The AMBIGUOUS phrase must be an exact substring of the FIGURATIVE and LITERAL sentences.
        \end{itemize}

        \\\midrule
        
        \multirow{1}{*}{\rotatebox[origin=c]{90}{\parbox[c]{4.7cm}{\large\textbf{In-Context Examples}}}} &\footnotesize
        $\begin{cases}
            \texttt{User: break the ice} \\
        \end{cases}$
        
        $\begin{cases}
            \texttt{Assistant: \{ } \\
            \texttt{\quad``AMBIGUOUS'': ``<broke the ice with our teammates>'',}\\
            \texttt{\quad``FIGURATIVE'': ``we <broke the ice with our teammates> by initiating a conversation'',}\\
            \texttt{\quad``LITERAL'': ``holding steel ice hammers tightly, we <broke the ice with our teammates>''}\\
            \texttt{\}}
        \end{cases}$
        $\begin{cases}
            \texttt{User: felt pins and needles} \\
        \end{cases}$
        
        $\begin{cases}
            \texttt{Assistant: \{ } \\
            \texttt{\quad``AMBIGUOUS'': ``<felt pins and needles>'',}\\
            \texttt{\quad``FIGURATIVE'': ``waiting for our boss to respond, we <felt pins and needles> in the room'',}\\
            \texttt{\quad``LITERAL'': ``she <felt pins and needles> on the floor when she spilled sewing supplies''}\\
            \texttt{\}}
        \end{cases}$

        \\\midrule

        \multirow{1}{*}{\rotatebox[origin=c]{90}{\parbox[c]{4.4cm}{\large\textbf{Added Messages}}}}&\footnotesize
        $\begin{cases}
        \texttt{User: goose egg} \\
        \end{cases}$

        $\begin{cases}
            \texttt{Assistant: \{ } \\
            \texttt{\quad``AMBIGUOUS'': ``<got a goose egg>'',}\\
            \texttt{\quad``FIGURATIVE'': ``he <\textcolor{red}{scored} a goose egg> after studying so diligently for the test'',}\\
            \texttt{\quad``LITERAL'': ``she <got a goose egg> from her grandfather's backyard''}\\
            \texttt{\}}
        \end{cases}$

        $\begin{cases}
        \texttt{\textcolor{red}{User: The AMBIGUOUS phrase must be an exact substring of the FIGURATIVE sentence, but yours}}\\
        \texttt{\textcolor{red}{isn't. Rewrite your AMBIGUOUS phrase, FIGURATIVE sentence, and LITERAL sentence accordingly.}}\\
        \end{cases}$

        $\begin{cases}
            \texttt{Assistant: \{ } \\
            \texttt{\quad``AMBIGUOUS'': ``<got a goose egg>'',}\\
            \texttt{\quad``FIGURATIVE'': ``he <got a goose egg> after studying so diligently for the test'',}\\
            \texttt{\quad``LITERAL'': ``she <got a goose egg> from her grandfather's backyard''}\\
            \texttt{\}}
        \end{cases}$
        
        \\\bottomrule
    \end{tabularx}
    \caption{\textbf{Prompt for \lm{GPT-4}} to generate one triple. The instruction is stated once, followed by two complete in-context examples. Finally, only the idiom is provided for the last example. In this case, the generation does not meet the requirement that the ambiguous subsentence must be a substring of the figurative sentence. We request a new triple, to which GPT responds with a triple that meets all the requirements.}
    \label{tab:generate_data}
\end{table*}

\end{document}